\documentclass[11pt,a4paper]{article}
\usepackage[hyperref]{emnlp2018}
\usepackage{times}
\usepackage{latexsym}

\usepackage{url}
\usepackage{amsmath}
\usepackage{algorithm}
\usepackage{algpseudocode}
\usepackage{graphicx}
\aclfinalcopy 
\def\aclpaperid{2016} 

\title{Hierarchical CVAE for Fine-Grained Hate Speech Classification}

\author{Jing Qian, Mai ElSherief, Elizabeth Belding, William Yang Wang\\
  Department of Computer Science \\
  University of California, Santa Barbara\\
  Santa Barbara, CA 93106 USA\\
  {\tt \{jing\_qian,mayelsherif,ebelding,william\}@cs.ucsb.edu} \\}

\date{}

\begin{document}
\maketitle
\begin{abstract}
Existing work on automated hate speech detection typically focuses on binary classification or on differentiating among a small set of categories. In this paper, we propose a novel method on a fine-grained hate speech classification task, which focuses on differentiating among 40 hate groups of 13 different hate group categories. We first explore the Conditional Variational Autoencoder (CVAE) ~\cite{larsen2016autoencoding,sohn2015learning} as a discriminative model and then extend it to a hierarchical architecture to utilize the additional hate category information for more accurate prediction. Experimentally, we show that incorporating the hate category information for training can significantly improve the classification performance and our proposed model outperforms commonly-used discriminative models.
\end{abstract}

\section{Introduction}
\label{sec:intro}
The impact of the vast quantities of user-generated web content can be both positive and negative. While it improves information accessibility, it also can facilitate the propagation of online harassment, such as hate speech. Recently, the Pew Research Center\footnote{http://www.pewinternet.org/2017/07/11/online-harassment-2017/} reported that ``roughly four-in-ten Americans have personally experienced online harassment, and 63\% consider it a major problem. Beyond the personal experience, two-thirds of Americans reported having witnessed abusive or harassing behavior towards others online.''
\begin{figure}[t]
\centering
\includegraphics[width=0.46\textwidth]{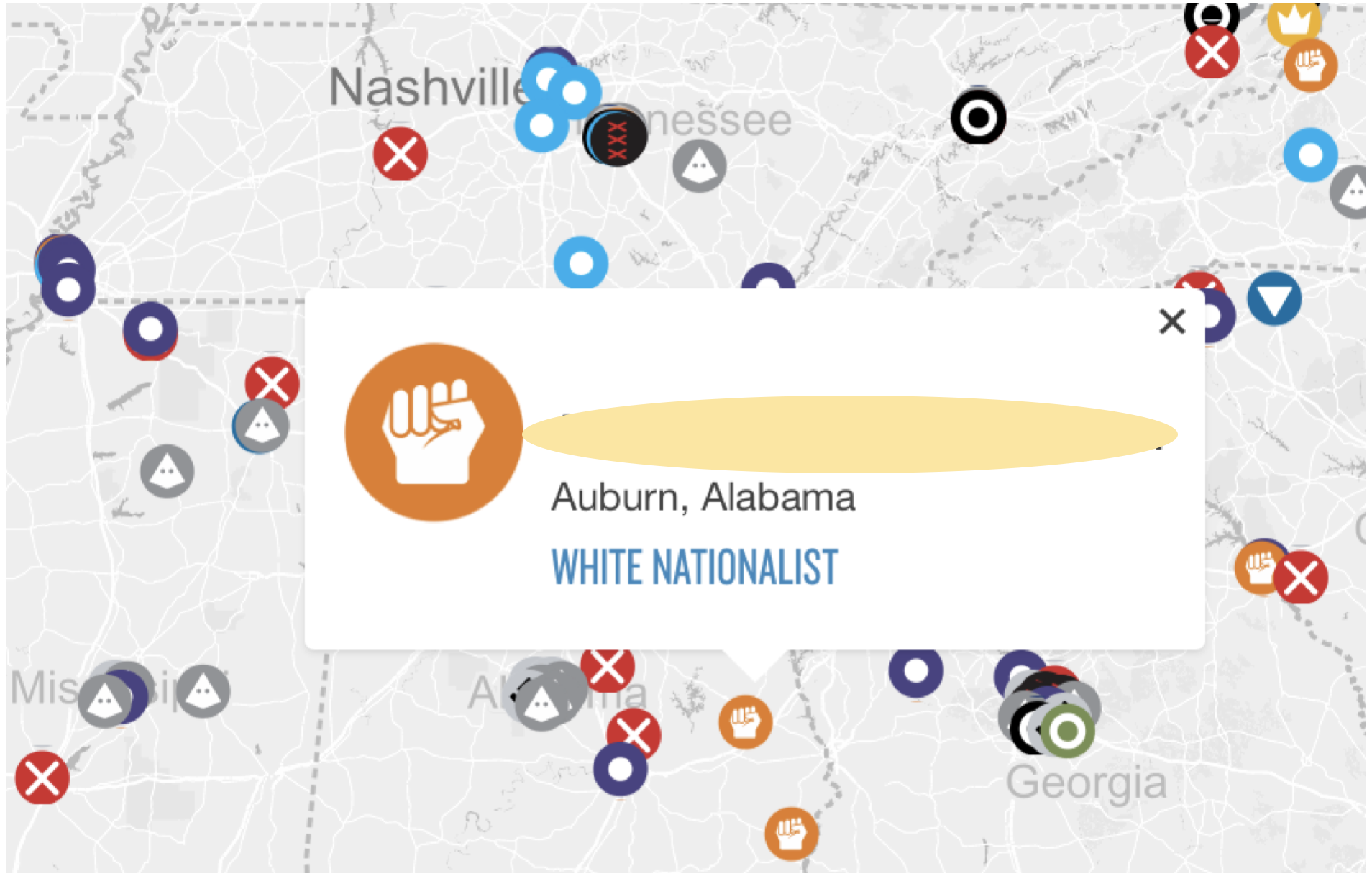}
\caption{A portion of the hate group map published by the Southern Poverty Law Center (SPLC). Each marker represents a hate group. The markers with the same pattern indicate the corresponding hate groups share the same ideology. The white box shows an example of a hate group in Auburn, Alabama under the category of "White Nationalist". Due to the sensitivity of the data, we mask the name of the group.}
\label{fig:hategroupmap}
\end{figure}

In response to the growth in online hate, there has been a trend of developing automatic hate speech detection models to alleviate online harassment~\cite{warner2012detecting,waseem2016hateful}. However, a common problem with these methods is that they focus on coarse-grained classifications with only a small set of categories. To the best of our knowledge, the existing work on hate speech detection formulates the task as a classification problem with no more than seven classes. Building a model for more fine-grained multiclass classification is more challenging since it requires the model to capture finer distinctions between each class. 

Moreover, fine-grained classification is necessary for fine-grained hate speech analysis. Figure~\ref{fig:hategroupmap} is a portion of the hate group map published by the Southern Poverty Law Center (SPLC)\footnote{https://www.splcenter.org}, where a hate group is defined as ``an organization that – based on its official statements or principles, the statements of its leaders, or its activities – has beliefs or practices that attack or malign an entire class of people, typically for their immutable characteristics.'' The SPLC divides the 954 hate groups in the United States into 17 categories according to their ideologies (e.g. Racist Skinhead, Anti-Immigrant, and others). The SPLC monitors hate groups throughout the United States by a variety of methodologies to determine the activities of groups and individuals: reviewing hate group publications and reports by citizens, law enforcement, field sources and the news media, and conducting their own investigations. Therefore, building automatic classification models to differentiate between the social media posts from different hate groups is both challenging and of practical significance.

In this paper, we propose a fine-grained hate speech classification task that separates tweets posted by 40 hate groups of 13 different hate group categories. Although CVAE is commonly used as a generative model, we find it can achieve competitive results and tends to be more robust when the size of the training dataset decreases, compared to the commonly used discriminative neural network models. Based on this insight, we design a Hierarchical CVAE model (HCVAE) for this task, which leverages the additional hate group category (ideology) information for training. 
Our contributions are three-fold:
\begin{itemize}
\itemsep=-2pt
\item  This is the first paper on fine-grained hate speech classification that attributes hate groups to individual tweets.
\item We propose a novel Hierarchical CVAE model for fine-grained tweet hate speech classification.
\item Our proposed model improves the Micro-F1 score of up to 10\% over the baselines.
\end{itemize}
In the next section, we outline the related work on hate speech detection, fine-grained text classification, and Variational Autoencoder. In Section~\ref{sec:cvae}, we explore the CVAE as a discriminative model, and our proposed method is described in Section~\ref{sec:our}. Experimental results are presented and discussed in Section~\ref{sec:experiments}. Finally, we conclude in Section~\ref{sec:conclusion}.

\section{Related Work}

\subsection{Hate Speech Detection}
An extensive body of work has been dedicated to automatic hate speech detection. Most of the work focuses on binary classification. \citet{warner2012detecting} differentiate between anti-semitic or not. \citet{gao2017recognizing}, \citet{zhong2016content}, and \citet{nobata2016abusive} differentiate between abusive or not. \citet{waseem2016hateful}, \citet{burnap2016us}, and \citet{davidson2017automated} focus on three-way classification. \citet{waseem2016hateful} classify each input tweet as racist hate speech, sexist hate speech, or neither. \citet{burnap2016us} build classifiers for hate speech based on race, sexual orientation or disability, while \citet{davidson2017automated} train a model to differentiate among three classes: containing hate speech, only offensive language, or neither.
\citet{badjatiya2017deep} use the dataset provided by \citet{waseem2016hateful} to do three-way classification. Our work is most closely related to \cite{van2015detection}, which focuses on fine-grained cyberbullying classification. However, this study only focuses on seven categories of cyberbullying while our dataset consists of 40 classes. Therefore, our classification task is much more fine-grained and challenging.  
 
\subsection{Fine-grained Text Classification}
Our work is also related to text classification. Convolutional Neural Networks (CNN) have been successfully applied to the text classification task. \citet{kim2014convolutional} applies CNN at the word level while \citet{zhang2015character} apply CNN at the character level. \citet{johnson2015effective} exploit the word order of text data with CNN for accurate text categorization. \citet{socher2013recursive} introduces the Recursive Neural Tensor Network for text classification. Recurrent Neural Networks (RNN) are also commonly used  for text classification \cite{tai2015improved,yogatama2017generative}. \citet{lai2015recurrent} and \citet{zhou2015c} further combine RNN with CNN. \citet{tang2015document} and \citet{yang2016hierarchical} exploit the hierarchical structure for document classification. \citet{tang2015document} generate from the sentence representation to the document representation while \citet{yang2016hierarchical} generate from the word-level attention to the sentence-level attention. However, the division of the hierarchies in our HCVAE is according to semantic levels, rather than according to document compositions. We generate from  the category-level representations to the group-level representations. Moreover, the most commonly used datasets by these works (Yelp reviews, Yahoo answers, AGNews, IMDB reviews~\cite{diao2014jointly}) have no more than 10 classes. 

\subsection{Variational Autoencoder}
Variational Autoencoder (VAE)~\cite{kingma2013auto} has achieved competitive results in many complicated generation tasks, such as handwritten digits~\cite{kingma2013auto,salimans2015markov}, faces~\cite{kingma2013auto,rezende2014stochastic}, and machine translation~\cite{zhang2016variational}. CVAE ~\cite{larsen2016autoencoding,sohn2015learning} is an extension of the original VAE framework that incorporates conditions during generation. In addition to image generation, CVAE has also been successfully applied to several NLP tasks, such as dialog generation ~\cite{zhao2017learning}. Although so far CVAE has always been used as a generative model, we explore the performance of the CVAE as a discriminative model and further propose a hierarchical CVAE model, which exploits the hate group category (ideology) information for training.

\section{CVAE Baseline}
\label{sec:cvae}
We formulate our classification task as the following equation:
\begin{equation}
Obj=\sum_{(x,y_g)\in X}\log p(y_g|x)
\end{equation}
where $x$, $y_g$ are the tweet text and hate group label respectively, $X$ is the dataset.
\begin{figure}[t]
\centering
 \includegraphics[width=0.45\linewidth]{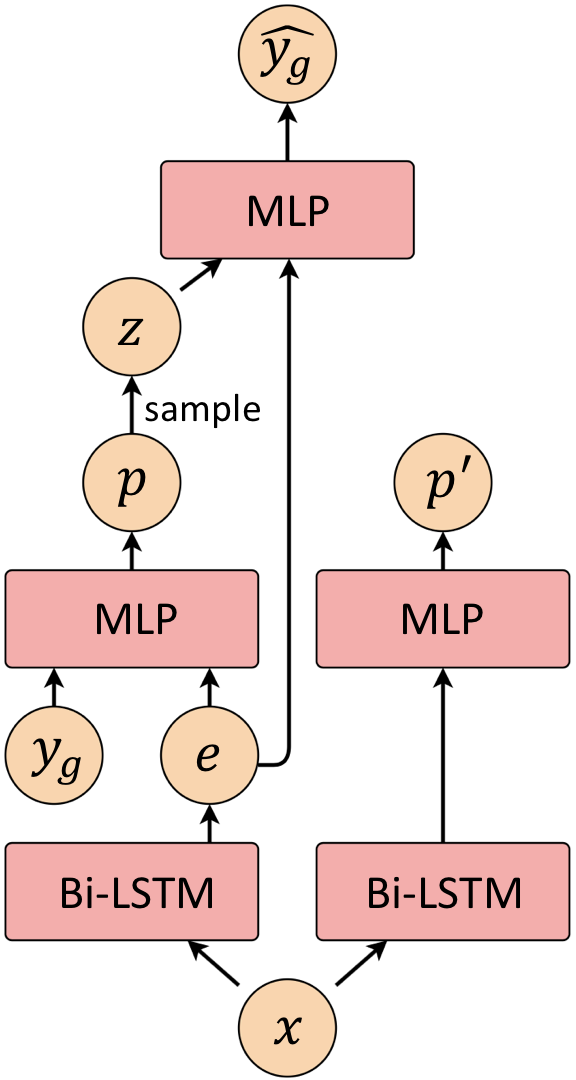}
\caption{The structure of the baseline CVAE model during training. Bi-LSTM is a bidirectional LSTM layer. MLP is a Multilayer Perceptron. $x$ is the embedded input text. $y_g$ is the ground truth hate group label. $\hat{y_g}$ is the output prediction of the hate group. $p$ is the posterior distribution of the latent variable $z$ while $p^\prime$ is the prior distribution of $z$. Note that this structure is used for training. During testing, the posterior network is replaced by the prior network to compute $\hat{y_g}$ and thus $y_g$ is not available during testing. Refer to Section~\ref{sec:cvae} for detailed explanation.}
\label{fig:baselinecvae}
\end{figure}
Instead of directly parameterizing $p(y_g|x)$, it can be further written as the following equation: 
\begin{equation}
p(y_g|x)=\int_{z}p(y_g|z,x)p(z|x)dz
\end{equation}
where $z$ is the latent variable. Since the integration over $z$ is intractable, we instead try to maximize the corresponding evidence lower bound (ELBO):
\begin{equation}
\begin{aligned}
ELBO=&E[\log p(y_g|z,x)]-\\
&D_{KL}[q(z|x,y_g)||p(z|x)]
\end{aligned}
\end{equation}
where $D_{KL}$ is the Kullback–Leibler (KL) divergence. $p(y_g|z,x)$ is the likelihood distribution, $q(z|x,y_g)$ is the posterior distribution, and $p(z|x)$ is the prior distribution. These three distributions are parameterized by $p_\varphi(y_g|z,x)$, $q_\alpha(z|x,y_g)$, and $p_\beta(z|x)$.
Therefore, the training loss function is:
\begin{equation}
\begin{aligned}
\mathcal{L}=&L_{REC}+L_{KL}\\=&E_{z\sim p_\alpha(z|x,y_g)}[-\log p_\varphi(y_g|z,x)]+\\
&D_{KL}[q_\alpha(z|x,y_g)||p_\beta(z|x)]
\end{aligned}
\end{equation}
The above loss function consists of two parts. The first part $L_{REC}$ is the reconstruction loss. Optimizing $L_{REC}$ can push the predictions made by the posterior network and the likelihood network closer to the ground truth labels. The second part $L_{KL}$ is the KL divergence loss. Optimizing it can push the output distribution of the prior network and that of the posterior network closer to each other, such that during testing, when the ground truth label $y_g$ is no longer available, the prior network can still output a reasonable probability distribution over $z$.

Figure~\ref{fig:baselinecvae} illustrates the structure of the CVAE model during training (the structure used for testing is different). The likelihood network $p_\varphi(y_g|z,x)$ is a Multilayer Perceptron (MLP). The structure of both the posterior network $q_\alpha(z|x,y_g)$ and the prior network $p_\beta(z|x)$ is a bidirectional Long Short-Term Memory (Bi-LSTM) ~\cite{hochreiter1997long} layer followed by an MLP. The Bi-LSTM is used to encode the tweet text. The only difference between the posterior and the prior network is that for the posterior network the input for the MLP is the encoded text concatenated with the group label while for the prior network only the encoded text is fed forward. During testing, the posterior network will be replaced by the prior network to generate the distribution over the latent variable (i.e. $p^\prime$ will replace $p$). Thus during testing, the ground truth labels will not be used to make predictions. 

We assume the latent variable $z$ has a multivariate Gaussian distribution: $p=\mathcal{N}(\mu, \Sigma)$ for the posterior network, and $p^\prime=\mathcal{N}(\mu^\prime, \Sigma^\prime)$ for the prior network. The detailed computation process is as follows:
\begin{align}
e&=f(x)\\
\mu,\Sigma&=s(y_g\oplus e)\\
\mu^\prime,\Sigma^\prime&=s^\prime(f^\prime(x))
\end{align}
where $f$ is the Bi-LSTM function and $e$ is the output of the Bi-LSTM layer at the last time step. $s$ is the function of the MLP in the posterior network and $s^\prime$ is that of the MLP in the prior network. The notation $\oplus$ means concatenation.
The latent variables $z$ and $z^\prime$ are randomly sampled from $\mathcal N(\mu,\Sigma)$ and $\mathcal N(\mu^\prime,\Sigma^\prime)$, respectively. During training, the input for the likelihood network is $z$:
\begin{equation}
\hat{y_g}=w(z)
\end{equation}
where $w$ is the function of the MLP in the likelihood network.
During testing, the prior network will substitute for the posterior network. Thus for testing, the input for the likelihood is $z^\prime$ instead of $z$: 
\begin{equation}
\hat{y_g}=w(z^\prime)
\end{equation}


\section{Our Approach}
\label{sec:our}
One problem with the above CVAE method is that it utilizes the group label for training, but ignores the available hate group category (ideology) information of the hate speech. As mentioned in Section~\ref{sec:intro}, hate groups can be divided into different categories in terms of ideologies. Each hate group belongs to a specific hate group category. Considering this hierarchical structure, the hate category information can potentially help the model to better capture the subtle differences between the hate speech from different hate groups. Therefore, we extend the baseline CVAE model to incorporate the category information. In this case, the objective function is as follows:
\begin{equation}
\begin{split}
Obj\!&=\!\!\!\sum_{(x,y_c,y_g)\in X}\!\!\!\!\log p(y_c,y_g|x)\\
&=\!\!\!\sum_{(x,y_c,y_g)\in X}\!\!\!\!\log p(y_c|x)\!+\!\log p(y_g|x,y_c)
\end{split}
\end{equation}
where $y_c$ is the hate group category label and 
\begin{align}
p(y_c|x)\!&=\!\!\int_{z_c}p(y_c|z_c,x)p(z_c|x)dz_c\\
p(y_g|x,y_c)\!&=\!\!\int_{z_g}\!\!\!p(y_g|z_g,\!x,\!y_c)p(z_g|x,\!y_c)dz_g
\end{align}
where $z_c$ and $z_g$ are latent variables. Therefore, the ELBO of our method is the sum of the ELBOs of $\log p(y_c|x)$ and $\log p(y_g|x,y_c)$:
\begin{equation}
\begin{aligned}
\!\!\!ELBO\!=&E[\log p(y_c|z_c,x)]-\\
&D_{KL}[q(z_c|x,y_c)||p(z_c|x)]+\\
&E[\log p(y_g|z_g,x,y_c)]-\\
&D_{KL}[q(z_g|x,y_c,y_g)||p(z_g|x,y_c)]
\end{aligned}
\end{equation}
During testing, the prior networks will substitute the posterior networks and the ground truth labels $y_c$ and $y_g$ are not utilized. Hence the prior $p(z_g|x,y_c)$ in the above equation cannot be parametrized as a network that directly takes the ground truth label $y_c$ and $x$ as inputs. Instead, we parameterize it as shown in the right part of Figure~\ref{fig:cvaebased}. We assume that $u^\prime$ is trained to be a latent representation of $y_c$, so we use $u^\prime$ and $x$ as inputs for this prior network. 

According to the ELBO above, the corresponding loss function $\mathcal{L}$ is the combination of the loss function for the category classification ($\mathcal{L}_c$) and that for the group classification ($\mathcal{L}_g$).
\begin{equation}
\begin{split}
\mathcal{L}=&\mathcal{L}_c+\mathcal{L}_g\text{, where}\\
\mathcal{L}_c=&E_{z_c\sim q_\alpha(z_c|x,y_c)}[-\log p_\varphi(y_c|z_c,x)]+\\
&D_{KL}[q_\alpha(z_c|x,y_c)||p_\beta(z_c|x)]\text{, and}\\
\mathcal{L}_g=&E_{z_g\sim q_\eta(z_g|x,y_c,y_g)}[-\log p_\theta(y_g|z_g,x,y_c)]+\\
&D_{KL}[q_\eta(z_g|x,y_c,y_g)||p_\gamma(z_g|x,u)]
\end{split}
\end{equation}
Figure~\ref{fig:cvaebased} shows the structure of our model for training. By assuming the latent variables $z_c$ and $z_g$ have multivariate Gaussian distributions, the actual outputs of the posterior and prior networks are the mean and variance: $p_c=\mathcal{N}(\mu_c, \Sigma_c)$, $p_g=\mathcal{N}(\mu_g, \Sigma_g)$ for the two posterior networks, and $p_c^\prime=\mathcal{N}(\mu_c^\prime, \Sigma_c^\prime)$, $p_g^\prime=\mathcal{N}(\mu_g^\prime, \Sigma_g^\prime)$ for the two prior networks. Note that in addition to these four distributions, there is another distribution $p_x=\mathcal{N}(\mu_x,\sigma_x)$ as shown in Figure~\ref{fig:cvaebased}. This distribution is generated only with the input text $x$, so it provides the basic distribution (in order to avoid confusion, we call it the basic distribution instead of the prior distribution) for both two posterior networks and two prior networks. With this basic distribution, the two posterior networks only need to capture the additional signals learned from $x$ and labels. Similarly, the prior networks only need to learn the additional signals learned by the posterior networks. The detailed computation process during training is shown as the following equations: 
\begin{figure}[t]
\centering
\includegraphics[width=0.97\linewidth]{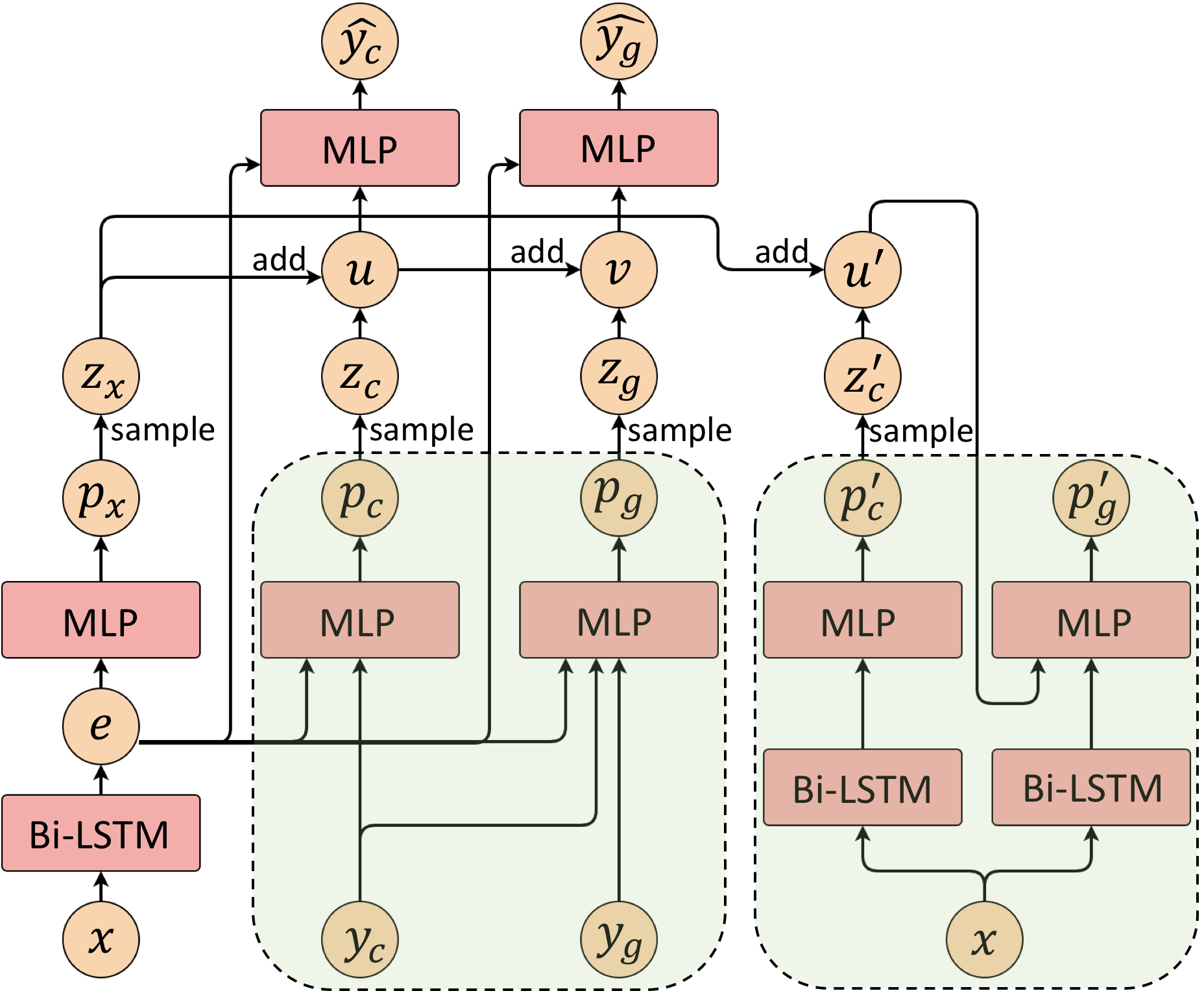}
\caption{The structure of the HCVAE during training. Bi-LSTM is a bidirectional LSTM layer. MLP is a Multilayer Perceptron. $x$ is the embedded input text. $y_c$ and $y_g$ are the ground truth hate category label and hate group label. $\hat{y_c}$ and $\hat{y_g}$ are the output predictions of the hate category and hate group. $z_c$, $z_g$, and $z_c^\prime$ are latent variables. In the left dotted box are two posterior networks. In the right dotted box are two prior networks. Note that this structure is used for training. During testing, the posterior networks will be substituted by the posterior networks (i.e. the left dotted box will be substituted with the right one) to compute $\hat{y_c}$ and $\hat{y_g}$. Thus $y_c$ and $y_g$ are not available during testing. Refer to Section~\ref{sec:our} for detailed explanation.}
\label{fig:cvaebased}
\end{figure}
\begin{algorithm}[t]
\small
  \caption{Train \& Test Algorithm}\label{algo:our}
  \begin{algorithmic}[1]
  \Function{train}{$X$}
    \State randomly initialize network parameters;
    \For{$epoch=1, E$}
      \For{$(text, category, group)$ in $X$}
        \State get embeddings $x$ and one-hot vectors $y_c$, $y_g$;
        \State compute $e$ with the Bi-LSTM;
        \State compute $\mu_x$, $\Sigma_x$, $\mu_c$, $\Sigma_c$, $\mu_g$, $\Sigma_g$;
        \State sample $z_x=reparameterize(\mu_x,\Sigma_x)$;
        \State sample $z_c=reparameterize(\mu_c,\Sigma_c)$;
        \State sample $z_g=reparameterize(\mu_g,\Sigma_g)$;
        \State $u=z_x+z_c$;
        \State $v=u+z_g$;
        \State compute $\hat{y_c}$ and $\hat{y_g}$ according to Eq.~\ref{eq:0}-~\ref{eq:2};
        \State $\mathcal{L}_{REC}=BCE(\hat{y_c},y_c)+BCE(\hat{y_g},y_g)$;
        \State compute $\mu_c^\prime$, $\Sigma_c^\prime$;
        \State sample $z_c^\prime=reparameterize(\mu_c^\prime,\Sigma_c^\prime)$;
        \State $u^\prime=z_x+z_c^\prime$;
        \State compute $\mu_g^\prime$, $\Sigma_g^\prime$;
        \State $\mathcal{L}_{KL}=D_{KL}(p_c||p_c^\prime)+D_{KL}(p_g||p_g^\prime)$;
        \State $\mathcal{L}=\mathcal{L}_{KL}+\mathcal{L}_{REC}$;
        \State update network parameters on $\mathcal{L}$; 
      \EndFor
    \EndFor
   \EndFunction
   \State
   \Function{test}{$X$}
   	\For{$text$ in $X$}
    	\State get embeddings $x$;
        \State compute $e$ with the Bi-LSTM;
        \State compute $\mu_x$, $\Sigma_x$, $\mu_c^\prime$, $\Sigma_c^\prime$;
        \State sample $z_x=reparameterize(\mu_x,\Sigma_x)$;
        \State sample $z_c^\prime=reparameterize(\mu_c^\prime,\Sigma_c^\prime)$;
        \State $u^\prime=z_x+z_c^\prime$;
        \State compute $\mu_g^\prime$, $\Sigma_g^\prime$;
        \State sample $z_g^\prime=reparameterize(\mu_g^\prime,\Sigma_g^\prime)$;
        \State $v^\prime=u^\prime+z_g^\prime$;
        \State compute $\hat{y_c}$ and $\hat{y_g}$ according to Eq.~\ref{eq:3}-~\ref{eq:4};
    \EndFor
   \EndFunction
  \end{algorithmic}
\end{algorithm}
\begin{figure*}[t]
\centering
\includegraphics[width=0.98\textwidth]{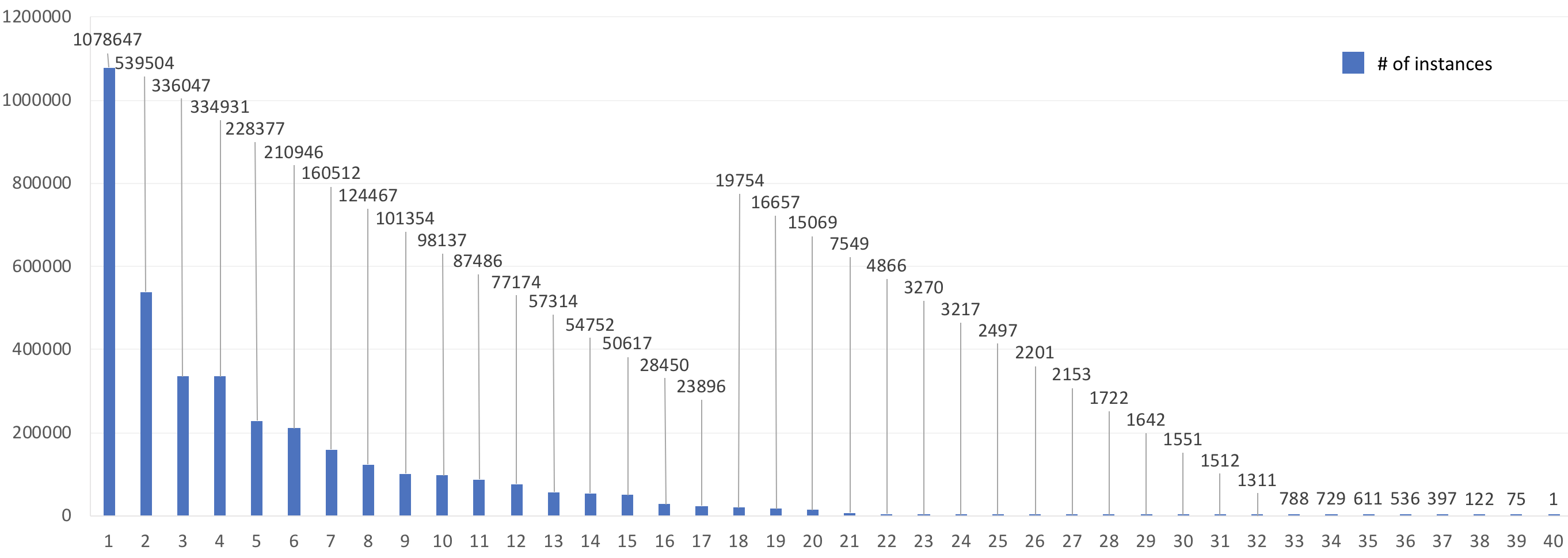}
\caption{The distribution of the data for 40 hate groups. The X-Axis is each hate group. The Y-Axis is the number of tweets collected for each hate group.
}
\label{fig:data}
\end{figure*}
\begin{align}
e&=f(x)\\
\mu_x,\Sigma_x&=s_x(e)\\
\mu_c,\Sigma_c&=s_c(y_c\oplus e)\\
\mu_g,\Sigma_g&=s_g(y_c\oplus y_g\oplus e)\\
\mu_c^\prime,\Sigma_c^\prime&=s_c^\prime(f_c^\prime(x))
\end{align}
where $f$ is the Bi-LSTM function and e is the output of the Bi-LSTM layer at the last time step. $s_x$, $s_c$, $s_g$, and $s_c^\prime$ are the functions of four different MLPs. $f_c^\prime$ is the Bi-LSTM function in the prior network $p_\beta(z_c|x)$.
The latent variables $z_x$, $z_c$, $z_g$, and $z_c^\prime$ are randomly sampled from the Gaussian distributions $\mathcal N(\mu_x,\Sigma_x)$, $\mathcal N(\mu_c,\Sigma_c)$, $\mathcal N(\mu_g,\Sigma_g)$, and $\mathcal N(\mu_c^\prime,\Sigma_c^\prime)$ respectively. As mentioned above, $p_x$ is the basic distribution while $p_c$, $p_g$, $p_c^\prime$, and $p_g^\prime$ are trained to capture the additional signals. Therefore, $z_x$ is added to $z_c$ and $z_c^\prime$, then the results $u$ and $u^\prime$ are further added to $z_g$ and $z_g^\prime$, respectively, as shown in the following equations:
\begin{align}
u^\prime&=z_x+z_c^\prime\\
\mu_g^\prime,\Sigma_g^\prime&=s_g^\prime(u^\prime \oplus f_g^\prime(x))\\
u&=z_x+z_c\label{eq:1}\\
v&=u+z_g
\end{align}
where $f_g^\prime$ is the Bi-LSTM function and $s_g^\prime$ is the function of the MLP in the prior network $p_\gamma(z_g|x,u)$. $+$ is element-wise addition.
During training, $u$ and $v$ are fed into the likelihood networks:
\begin{align}
\hat{y_c}&=w_c(e\oplus u)\label{eq:0}\\ 
\hat{y_g}&=w_g(e\oplus v) \label{eq:2}
\end{align}
where $w_c$ and $w_g$ are the functions of the MLPs in two likelihood networks. 
During testing, the prior networks will substitute the posterior networks, so the latent variable $z_g^\prime$ is randomly sampled from the Gaussian distributions $\mathcal N(\mu_g^\prime,\Sigma_g^\prime)$, and the last four equations above (Equation~\ref{eq:1} -~\ref{eq:2}) will be replaced by the following ones: 
\begin{align}
\hat{y_c}&=w_c(e\oplus u^\prime)\label{eq:3}\\
v^\prime&=u^\prime+z_g^\prime\\ 
\hat{y_g}&=w_g(e\oplus v^\prime)\label{eq:4}
\end{align}
Algorithm~\ref{algo:our} illustrates the complete training and testing process. $BCE$ refers to the Binary Cross Entropy loss. Note that during training, the ground truth category labels and group labels are fed to the posterior network to generate latent variables. But during testing, the latent variables are generated by the prior network, which only utilizes the texts as inputs.

\section{Experiments}
\label{sec:experiments}
\begin{table*}[t!]
\centering
\small
\begin{tabular}{|l|c|c|c|c|c|c|}
  \hline
  Dataset & \multicolumn{3}{|c|}{Complete} &\multicolumn{3}{|c|}{Subset}\\
  \hline
  Metric & Macro-F1 & Micro-F1 & Weighted-F1 &Macro-F1 &Micro-F1 &Weighted-F1 \\
  \hline
  \hline
  upper bound &.697 &.904 &.881 & -- & -- & -- \\
  \hline
  \hline
  SVM + tf-idf &.653 &.834 &.835 &\textbf{.521} &.771 &.772 \\
  \hline
  LR + tf-idf &.586 &.787 &.792 &.494 &.727 &.736 \\
  \hline
  \hline
   Char-CNN  &.604 &.840 &.853 &.457 &{.730} &.744 \\
  \hline
  Bi-LSTM &{.627} &.767 &.745 &.353 &.599 &.570 \\
  \hline
  CVAE &.520 &.799 &.830 &.453 &.774 &.784\\
  \hline
  HCVAE (our) &\textbf{.664} &\textbf{.844} &\textbf{.858} &{.469} &\textbf{.787} &\textbf{.799} \\
  \hline
\end{tabular}
\caption{Experimental results. Complete: The performance achieved when 90\% of the entire dataset is used for training. Subset: The performance achieved when only 10\% of the dataset is used for training. The best results are in bold. }
\label{tab:results}
\end{table*}
\subsection{Dataset}
We collect the data from 40 hate group Twitter accounts of 13 different hate ideologies, e.g., white nationalist, anti-immigrant, racist skinhead, among others. The detailed themes and core values behind each hate ideology are discussed in the SPLC ideology section.\footnote{https://www.splcenter.org/fighting-hate/extremist-files/ideology} For each hate ideology, we collect a set of Twitter handles based on hate groups identified by the SPLC center.\footnote{https://www.splcenter.org/fighting-hate/extremist-files/groups}
For each hate ideology, we select the top three handles in terms of the number of followers. Due to ties, there are four different groups in several categories of our dataset. The dataset consists of all the content (tweets, retweets, and replies) posted with each account from the group's inception date, as early as 07-2009, until 10-2017. Each instance in the dataset is a tuple of (tweet text, hate category label, hate group label). The complete dataset consists of approximately 3.5 million tweets. Note that due to the sensitive nature of the data,  we anonymously reference the Twitter handles for each hate group by using IDs throughout this paper. The distribution of the data is illustrated in Figure~\ref{fig:data}. 

\subsection{Experimental Settings}
In addition to the discriminative CVAE model described in Section~\ref{sec:cvae}, we implement for other baseline methods and an upper bound model as follows. \\
\textbf{Support Vector Machine (SVM)}: We implement an SVM model with linear kernels. We use L2 regularization and the coefficient is 1. The input features are the Term Frequency Inverse Document Frequency (TF-IDF) values of up to 2-grams.
\\
\textbf{Logistic Regression (LR)}: We implement the Logistic Regression model with L2 regularization. The penalty parameter C is set to 1. Similar to the SVM model, we use TF-IDF values of up to 2-grams as features.
\\
\textbf{Character-level Convolutional Neural Network (Char-CNN)}: We implement the character-level CNN model for text classification as described in ~\cite{zhang2015character}. It is 9 layers deep with 6 convolutional layers and 3 fully-connected layers. The configurations of the convolutional layers are kept the same as those in ~\cite{zhang2015character}.
\\
\textbf{Bi-LSTM}:
The model consists of a bidirectional LSTM layer followed by a linear layer. The embedded tweet text $x$ is fed into the LSTM layer and the output at the last time step is then fed into the linear layer to predict $\hat{y_g}$.
\\ 
\textbf{Upper Bound}: 
The upper bound model also consists of a bidirectional LSTM layer followed by a linear layer. The difference between this model and Bi-LSTM is that it takes the tweet text $x$ along with the ground truth category label $y_c$ as input during both training and testing. The LSTM layer is used to encode $x$. The encoding result is concatenated with the ground truth category label and then fed into the linear layer to give the prediction of the hate group $\hat{y_g}$. Since it utilizes $y_c$ for prediction during testing, this model sets an upper bound performance for our method.

For the baseline CVAE, Bi-LSTM, the upper bound model, and the HCVAE, we use randomly initialized word embeddings with size 200. All the neural network models are optimized by the Adam optimizer with learning rate 1e-4. The batch size is 20. The hidden size of all the LSTM layers in these models is 64 and all the MLPs are three-layered with non-linear activation function Tanh. For the CVAE and the HCVAE, the size of the latent variables is set to 256.

All the baseline models and our model are evaluated on two datasets. We first use the complete dataset for training and testing. 90\% of the instances are used for training and the rest for testing. In order to evaluate the robustness of the baseline models and our model, we also randomly selected a subset (10\%) of the original dataset for training while the testing dataset is fixed. Since the upper bound model is used to set an upper bound on performance, we do not evaluate it on the smaller training dataset. We use Macro-F1, Micro-F1, and Weighted-F1 to evaluate the classification results. As shown in Figure~\ref{fig:data}, the dataset is highly imbalanced, which causes problems for training the above models. In order to alleviate this problem, we use a weighted random sampler to sample instances from the training data. However, the testing dataset is not sampled, so the distribution of the testing dataset remains the same as that of the original dataset. This allows us to evaluate the models' performance on the data with a realistic distribution.

\subsection{Experimental Results}
The experimental results are shown in Table~\ref{tab:results}. The testing dataset keeps the imbalanced distribution of the original data, so the Macro-F1 score of each method is significantly lower than the Micro-F1 score and the Weighted-F1 score. Comparing the performance of the Bi-LSTM model and that of the baseline CVAE model shows that although CVAE is traditionally used as a generative model, it can also achieve competitive performance as a discriminative model. 
\begin{figure}[t]
\centering
\includegraphics[width=0.45\textwidth]{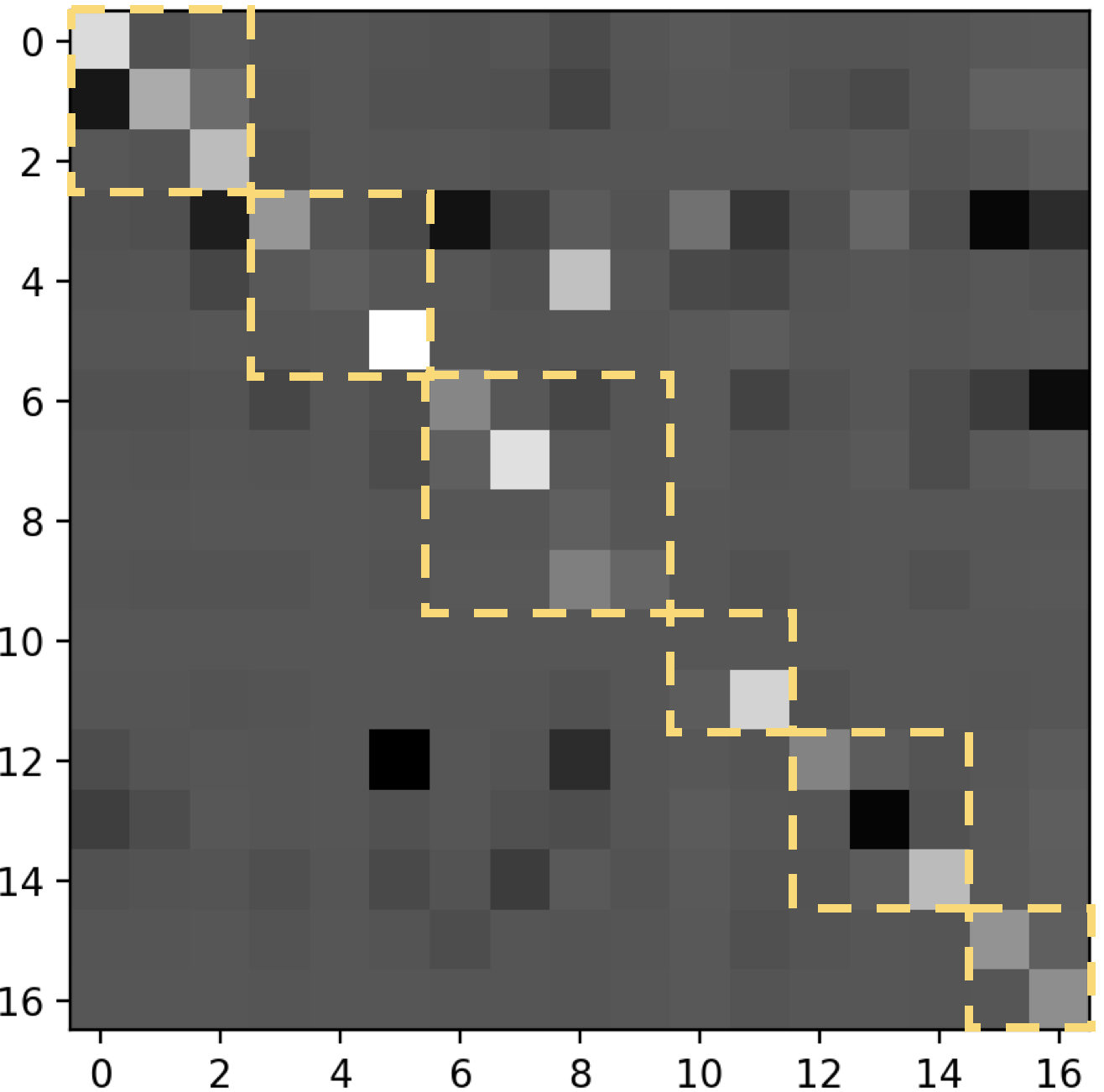}
\caption{This figure compares a subset (17 hate groups of 6 categories) of the predictions made by the baseline CVAE and the HCVAE. The X and Y axes are the index of each hate group. The hate groups of the same category are grouped together as shown in the dashed squares. The color of the grid $(i,j)$ is mapped from $r_{hcvae}(i,j)-r_{cvae}(i,j)$, where $r(i,j)$ is the fraction of the group $i$'s instances among the instances predicted as the group $j$. A higher difference value corresponds to a lighter color. A grid darker than the background is mapped from a negative value while a grid lighter than the background is mapped from a positive one.} 
\label{fig:prec}
\end{figure}
All the methods achieve lower F1 scores when using the smaller training dataset. Due to the imbalanced distribution of our dataset, half of the 40 groups has less than 1k tweets in the smaller training dataset, which leads to the sharp decline in the Macro-F1 scores of all the methods. However, the performance of both CVAE-based models degrades less than that of the other two neural network models (the Bi-LSTM model and the Char-CNN model) according to the Micro-F1 Weighted-F1 scores. These two CVAE-based models tend to be more robust when the size of the training dataset is smaller. The difference between the CVAE-based models and the other two models is that both the Bi-LSTM model and the Char-CNN model directly compress the input text into a fixed-length latent variable while the CVAE model explicitly models the posterior and likelihood distributions over the latent variable. The result is that, during testing, the inference strategy of the Bi-LSTM model and the Char-CNN model is actually comparing the compressed text to the compressed instances in the training dataset while the strategy of the CVAE-based models is to compare the prior distributions over the latent variable. Compared with the compressed text, prior distributions may capture higher level semantic information and thus enable better generalization. 

By further utilizing the hate category label for training, the HCVAE outperforms the baseline CVAE on all three metrics. Figure~\ref{fig:prec} illustrates the difference between the predictions made by the HCVAE and the CVAE. As shown in the figure, most of the lighter girds are in the dashed squares while most of the darker girds are outside. This indicates that the HCVAE model tends to correct the prediction of the CVAE's misclassified instances to be closer to the ground truth. In most cases, the misclassified instances can be directly corrected to the ground truth (the diagonal). In other cases, they are not corrected to the ground truth but are corrected to the hate groups of the same categories as the ground truth (the dashed squares). This shows that additional ideology information is useful for the model to better capture the subtle differences among tweets posted by different hate groups.  

Although our method outperforms the baseline methods, there is still a gap between its performance and the upper bound performance. We analyze this in the following section.

\subsection{Error Analysis}
\begin{figure}[t]
\centering
\includegraphics[width=0.5\textwidth]{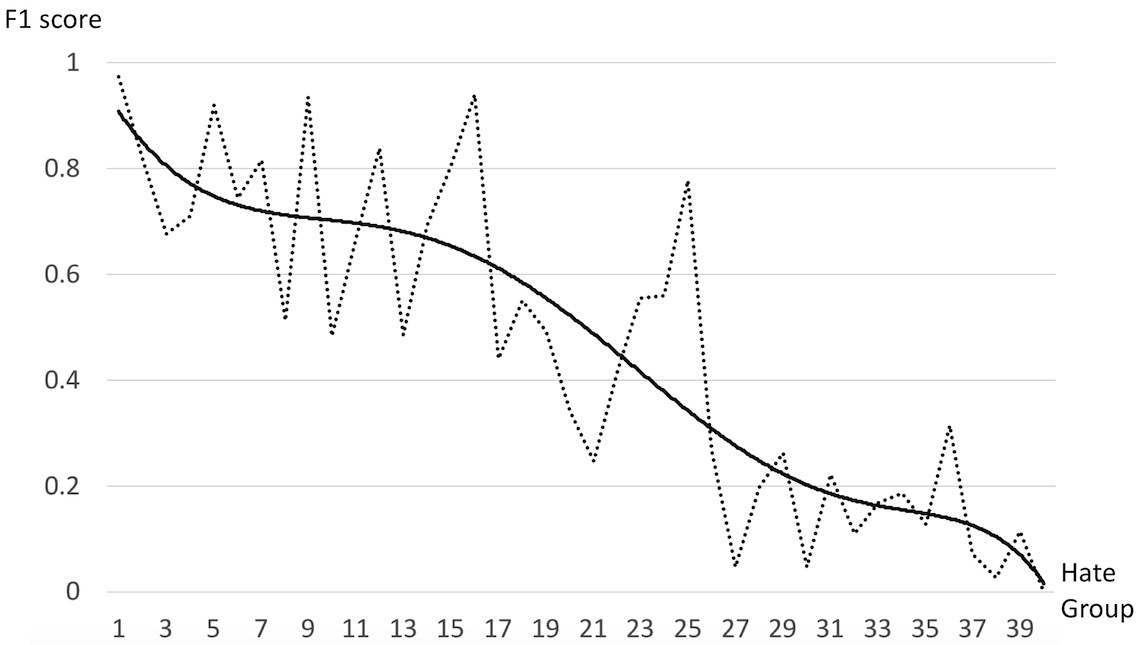}
\caption{The F1 score achieved by our method on each hate group. The Y-axis is the F1 score. The X-axis is the hate group sorted by the number of instances in the dataset from larger to smaller. The dashed line shows the F1 scores and the solid line is the corresponding trendline.
}
\label{fig:error}
\end{figure}
As mentioned above, in some cases our model misclassified the tweet as a group under the same category as the ground truth. Take, for instance, the tweet from \textit{Group1}: \textit{The only good \#muslim is a \#dead \#muslim}. Our HCVAE model predicts it as from \textit{Group2}. But both \textit{Group1} and \textit{Group2} are under the category ``neo nazi''.
One possible reason for this kind of error is that the imbalance of the dataset adversely affects the performance of our method. Figure~\ref{fig:error} shows the F1 scores achieved by the HCVAE on each group. The performance of the model tends to be much lower when the number of the group's training instances decreases.  Although we use the weighted random sampler to alleviate the problem of the imbalanced dataset, repeating the very limited data instances (less than 3k) of a group during training cannot really help the posterior and prior networks to give a reasonable distribution that can generalize well on unseen inputs. Therefore, when the model comes into the instances in the testing dataset, the performance can be less satisfactory. This is a common problem for all the methods, which is the cause of the significantly lower Macro-F1 scores.

Another type of error is caused by the noisy dataset. Take, for instance, the tweet from \textit{Group3} under the category ``ku klux klan'': \textit{we must secure the existence of our people and future for the White Children !!} Our model classifies it as from \textit{Group4} under the category ``neo nazi'', which makes sense but is an error.

\section{Conclusion}
\label{sec:conclusion}
In this paper, we explore the CVAE as a discriminative model and find that it can achieve competitive results. In addition, the performance of the CVAE-based models tend to be more stable than that of the others when the dataset gets smaller. We further propose an extension of the basic discriminative CVAE model to incorporate the hate group category (ideology) information for training. Our proposed model has a hierarchical structure that mirrors the hierarchical structure of the hate groups and the ideologies. We apply the HCVAE to the task of fine-grained hate speech classification, but this Hierarchical CVAE framework can be easily applied to other tasks where the hierarchical structure of the data can be utilized. 

\bibliography{emnlp2018}
\bibliographystyle{acl_natbib_nourl}

\appendix

\end{document}